\def\year{2025}\relax
\documentclass[letterpaper]{article} 
\usepackage{aaai25}  
\usepackage{times}  
\usepackage{helvet} 
\usepackage{courier}  
\usepackage[hyphens]{url}  
\usepackage{graphicx} 
\usepackage{natbib}  
\usepackage{caption} 
\usepackage{url}            
\usepackage{booktabs}       
\usepackage{amsfonts}       
\usepackage{nicefrac}       
\usepackage{microtype}      
\usepackage{xcolor}         
\usepackage{makecell}
\usepackage{graphicx}
\usepackage{soul}
\usepackage{amsmath, amsthm, amssymb}
\usepackage[most]{tcolorbox}
\usepackage{paralist}
\usepackage{bm}
\usepackage[american]{babel}
\usepackage{microtype}
\usepackage{latexsym}
\usepackage{booktabs}
\usepackage{multirow}
\usepackage{array}
\usepackage{natbib}
\usepackage{appendix}
\usepackage{algpseudocode} 
\usepackage{amssymb}
\usepackage{colortbl}
\usepackage{caption}
\usepackage{subcaption}
\usepackage{thmtools}
\usepackage{thm-restate}
\usepackage{cleveref}
\usepackage{multicol}

\definecolor{lightblue}{HTML}{328bb7}
\definecolor{lightred}{HTML}{d7907b}

\newtheorem{theorem}{Theorem}

\usepackage{mathtools}

\title{Sparse Orthogonal Parameters Tuning for Continual Learning}
\author{
    Kun-Peng Ning$^{*}$,
    Hai-Jian Ke$^{*}$,
    Yu-Yang Liu,
    Jia-Yu Yao,
    Yong-Hong Tian,
    Li Yuan$^{\dagger}$\\
}
\affiliations{
    \textit{\{ningkp,kehaijian\}@stu.pku.edu.cn, \{jiayu\_yao,yhtian,yuanli-ece\}@pku.edu.cn, liuyuyang@sia.cn}\\
    School of Electronic and Computer Engineering \\ 
    Peking University, Shenzhen, China \\
}

\begin{document}

\maketitle

\renewcommand{\thefootnote}{\relax}
\footnote{$^{*}$These authors contributed equally to this work.}
\footnote{$^{\dagger}$Corresponding authors.}
\renewcommand{\thefootnote}{\arabic{footnote}}

	
	

\begin{abstract}
Continual learning methods based on pre-trained models (PTM) have recently gained attention which adapt to successive downstream tasks without catastrophic forgetting. These methods typically refrain from updating the pre-trained parameters and instead employ additional adapters, prompts, and classifiers. In this paper, we from a novel perspective investigate the benefit of sparse orthogonal parameters for continual learning. We found that merging sparse orthogonality of models learned from multiple streaming tasks has great potential in addressing catastrophic forgetting. 
Leveraging this insight, we propose a novel yet effective method called SoTU (\textbf{S}parse \textbf{O}rthogonal Parameters \textbf{TU}ning).
We hypothesize that the effectiveness of SoTU lies in the transformation of knowledge learned from multiple domains into the fusion of orthogonal delta parameters. Experimental evaluations on diverse CL benchmarks demonstrate the effectiveness of the proposed approach. Notably, SoTU achieves optimal feature representation for streaming data without necessitating complex classifier designs, making it a Plug-and-Play solution.
\end{abstract}


\section{Introduction}
Continual learning (CL) has emerged as a fundamental challenge in the field of deep learning, aiming to enable models to continuously learn from new tasks while maintaining performance on previously learned knowledge \cite{incremental_survey,lifelong_machine}.
Traditional fully-supervised training paradigms face limitations in adapting to new scenarios because parameter updates become biased toward newer data, overwriting previously acquired knowledge. This phenomenon is often characterized as \textit{catastrophic forgetting} \cite{catastrophic_forgetting}. In response to this challenge, classical CL strategies range from rehearsal methods~\cite{rebuffi2017icarl} to regularization approaches~\cite{EWC,LWF} and dynamic expansion mechanisms~\cite{DEN}. However, most classical methodologies predominantly revolve around models that are ``trained from scratch'', commencing with randomly initialized parameters.

With the rapid development of pre-training techniques, continual learning methods~\cite{l2p,ranpac} based on pre-trained models (PTMs) have emerged. These methods capitalize on the robust generalization ability of PTMs derived from extensive datasets for various downstream tasks. Typically, these methods choose not to update pre-trained parameters and opt instead for the use of additional adapters, \cite{zhou2023revisiting,ranpac}, prompts \cite{l2p,dualprompt,codaprompt}, and classifiers \cite{zhou2022model,yan2021dynamically,wang2022foster}. 
Despite their success, some of them have complex structural designs that make it hard to implement in real-world applications.

\begin{figure*}[t]
    \centering
    \includegraphics[width=1\textwidth]{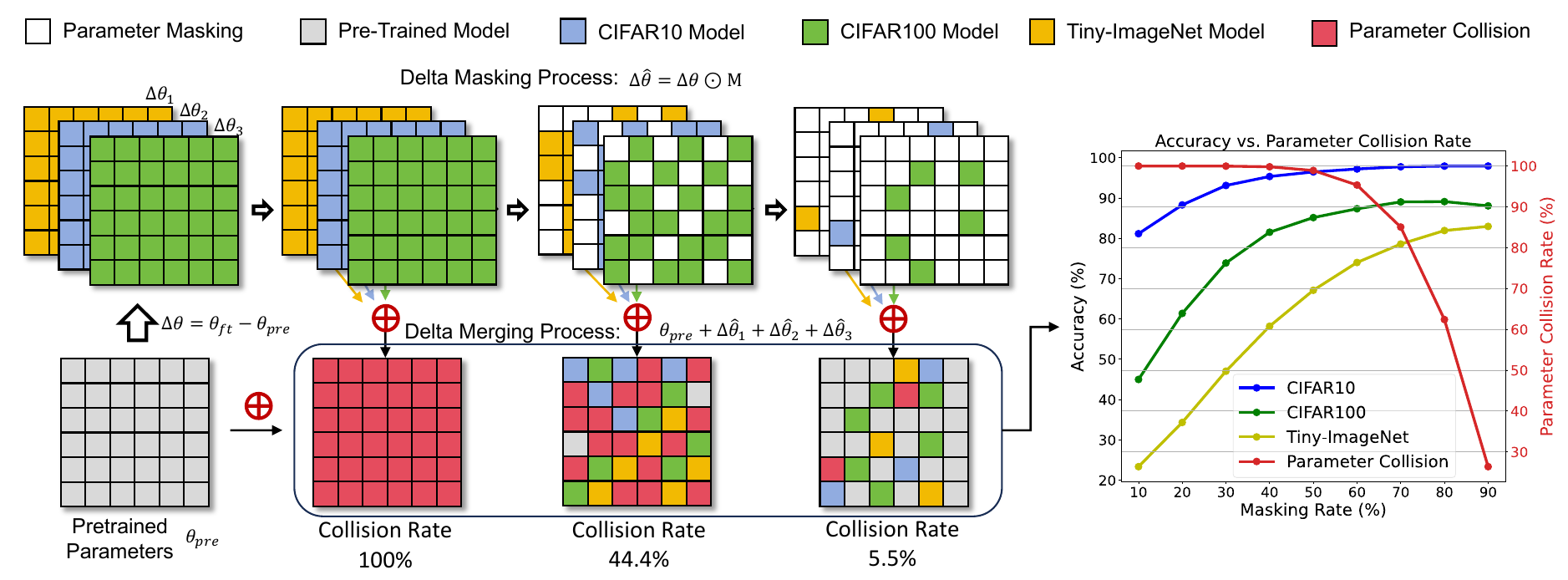}
    \caption{The \textit{sparse orthogonal} characteristic of pre-trained parameters. We randomly sample delta parameters and merge them from multiple domains into one feature extractor with different masking rates. We show that merging high-sparsity deltas can maintain comparable or even superior performance on all three tasks, while low sparsity will cause seriously parameter collisions, resulting in poor performance (right of figure). }
    \label{fig:1}
\end{figure*}

In this paper, we from a novel perspective investigate the characteristics of sparse orthogonal parameters for continual learning. As illustrated in Figure~\ref{fig:1}, we first fine-tuned three models ($\theta_{ft}$) based on the pre-trained ViT ($\theta_{pre}$) in CIFAR10, CIFAR100, and Tiny-ImageNet, respectively. Then, we calculate the delta parameters ($\Delta\theta=\theta_{ft}-\theta_{pre}$) for each task while randomly masking them with varying ratios based on the Bernoulli distribution ($\Delta\hat{\theta}=\Delta\theta\odot\boldsymbol{m}$). Next, when we merged these high-sparsity deltas from three tasks into one feature extractor, we surprisingly found that it achieved comparable or even superior performance on all three tasks. 
It is worth noting that merging delta parameters requires high sparsity, while low sparsity will cause serious parameter collisions, resulting in poor performance. 
As shown in the right of Figure \ref{fig:1}, when the masking rate is set to 10\%, the parameter conflicts can reach up to 99.96\%, and the model performance after delta merging process is extremely poor on these three datasets. On the other hand, as the masking rate increases, the parameter conflicts decrease and the model performance significantly improves. 
Therefore, we believe that merging sparse orthogonal delta parameters holds enormous promise in mitigating catastrophic forgetting problems.

Based on the findings above, we propose a novel yet effective CL approach called \textbf{SoTU} (\textbf{S}parse \textbf{O}rthogonal Parameters \textbf{TU}ning). The proposed SoTU encourages sparsity in updating parameters to decouple continual deltas and merge each other, while preventing the model from forgetting previously learned knowledge when adapting to new tasks. 
Specifically, to achieve sparse deltas, we multiply the original $\Delta\theta$ by a random binary mask matrix generated from the Bernoulli distribution. 
By controlling the sparsity level, we aim to approximate orthogonality among the delta matrices, thereby minimizing interference between delta updates and promoting independence of parameter changes. 
Additionally, we employ a streamlined merging algorithm to consolidate deltas from multiple fine-tuned models into a single cohesive model, further enhancing effectiveness and efficiency.

We extensively evaluate the proposed SoTU approach on six continual learning benchmarks, comparing it with twelve state-of-the-art baselines. The experimental results demonstrate that SoTU can effectively mitigate catastrophic forgetting problems, and achieve higher classification performance compared to existing CL methods. Notably, our method is noteworthy for its ability to achieve optimal feature representation for streaming data without the need for any elaborate classifier designs.

The contributions of this paper can be summarized as follows:
\begin{itemize}
    \item We found that merging multiple high-sparsity delta parameters can achieve competitive or even better performance on downstream tasks, which has great promise in combating catastrophic forgetting problems. 
    \item We demonstrated that the effectiveness behind the high-sparsity delta merging is due to the parameters' orthogonality across multiple tasks.
    \item Based on the \textit{sparse orthogonal} characteristic, we propose a novel CL method called SoTU that can achieve optimal feature representation for streaming data without the need for any elaborate classifier designs.
    \item Experimental results on multiple benchmarks demonstrate that the proposed approach can achieve higher classification performance than existing SoTU methods, especially in feature representation. 
\end{itemize}

\begin{figure*}[t]
    \centering
    \includegraphics[width=0.9\textwidth]{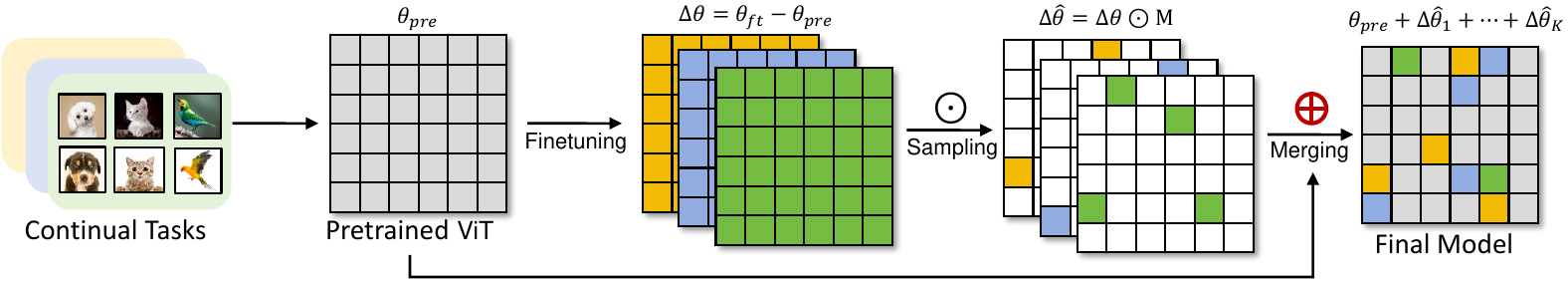}
    \caption{The framework of SoTU. Facing the continual tasks, we first fine-tune the pre-trained ViT and obtain the task-specific delta parameter $\Delta\theta$. Then, we randomly mask the delta with the Bernoulli distribution, while obtaining the high-sparsity delta $\Delta\hat{\theta}$. Finally, we can obtain the final mixture model $\theta^k$ by merging those high-sparsity deltas for all seen tasks.  }
    \vspace{-0.3cm}
    \label{fig:2}
\end{figure*}

\section{Related Work}
\textbf{Continual Learning (CL).} CL aims to adapt and learn new knowledge over time from a stream of continuous data. The key challenge in CL is catastrophic forgetting~\cite{catastrophic_forgetting}, where the model forgets previously learned knowledge when learning new tasks. To address this challenge, various methods \cite{EWC,icarl,LWF,liu2021l3doc,zhou2022model} have been proposed. Classical CL methods can be broadly classified into three categories: \emph{rehearsal}, \emph{regularization}, and \emph{dynamic expansion}. Rehearsal methods \cite{rebuffi2017icarl, zhou2022model}, store and replay examples from past tasks to reduce forgetting. Regularization approaches \cite{EWC,LWF} use terms to constrain updates for old data. Dynamic expansion \cite{DEN,yan2021dynamically} enlarges the model's capacity for new tasks while retaining old knowledge. 
Recently, pre-trained models (PTMs) have emerged for CL. These models, trained on large datasets, exhibit strong generalization and adaptability~\cite{zhou2024continual}. PTM-based CL approaches are categorized as prompt-based and representation-based.
Prompt-based methods, \textit{e.g.}, L2P \cite{l2p}, DualPrompt \cite{dualprompt}, and CODA-Prompt \cite{codaprompt}, propose to learn some task-specific prompts to adapt different downstream tasks. Representation-based methods (\textit{e.g.}, SimpleCIL \cite{janson2022simple}, ADAM \cite{zhou2023revisiting}, and RanPAC \cite{ranpac}) attempt to directly employ the inherent ability of PTM for classification with the nearest class mean (NCM) mechanism \cite{janson2022simple,pelosin2022simpler}. 
However, PTM-based CL still faces challenges like catastrophic forgetting and memory-intensive exemplar storage. 

\textbf{Model Mixture.} 
Another line of related work considers model mixture, which aims to create a set of models during the continual learning process and conduct model ensemble or model merge during inference \cite{zhou2024continual,wang2023isolation,wang2024hierarchical,LAE}. Among them, ESN \cite{wang2023isolation} creates a set of classifiers individually based on the same PTM during the learning process, \textit{i.e.}, it initializes and trains a new classifier head when facing a new task. During inference, it designs a voting strategy for these classifier heads by adopting a set of temperatures. LAE \cite{LAE} adopts a similar inference strategy by choosing the max logit across different models. HiDe-Prompt \cite{wang2024hierarchical} applies prompt merge after each continual learning stage, which combines multiple distinct models into a single unified model without requiring additional training.

\section{Method}
In this section, we first introduce the continual learning with PTMs approaches. Then, we introduce the proposed SoTU approach in detail, followed by the theoretical analysis. 

\subsection{Continual Learning with PTMs}
Continual learning \cite{de2021continual,mai2022online,zhou2023deep} focuses on the learning scenario involving a sequence of tasks $\{\mathcal{D}^1,\mathcal{D}^2,...,\mathcal{D}^T\}$. The $k$-th dataset $\mathcal{D}^k$ contains the set of input instances and their labels, \textit{i.e.}, $\mathcal{D}^k=\{(\boldsymbol{x}_i,\boldsymbol{y}_i)\}_{i=1}^{n_k}$ . Among them, $x_i \in \mathbb{R}^D$ is an instance of class $y_i\in\mathcal{Y}_k$, and $\mathcal{Y}_k$ is the label space of task $k$. During the current $k$-th training stage, we can only access data from $\mathcal{D}^k$. The goal is to continually learn a model $f$ for all seen tasks without forgetting, \textit{i.e.}, to minimize the following expected risk: 
\begin{equation}
f^*=\mathop{\arg\min}_{f\in\mathcal{H}}\, \, \mathbb{E}_{(\boldsymbol{x},\boldsymbol{y})\sim \mathcal{D}^1\cup\mathcal{D}^2\cup,...,\cup\mathcal{D}^k} [\ell(f(\boldsymbol{x}),\boldsymbol{y})],
\end{equation}
where $\mathcal{H}$ is the hypothesis space, and $\ell(\cdot)$ is the cross-entropy loss. For convenience, we decompose the classification model $f_\theta(\boldsymbol{x})$ into two parts: $f(\boldsymbol{x})=W^\intercal\phi_\theta(\boldsymbol{x})$, where $\phi_\theta(\cdot)$ is the embedding function parameterized by $\theta$, and $W$ is the classification head.

Instead of ``training from scratch'', many CL methods based on pre-trained models (PTM) have recently been proposed due to their powerful and generalizable feature extractor $\phi_{\theta_{pre}}(\cdot)$ \cite{l2p,ranpac,zhou2023revisiting}. Among them, most methods utilize an ImageNet21K \cite{deng2009imagenet} pre-trained Vision Transformer (ViT) \cite{dosovitskiy2020image} as an embedding function. They typically choose not to update pre-trained parameters $\theta_{pre}$ and instead employ additional adapters, prompts, and classifiers. For example, the \textit{prompt-based} methods aim to learn task-specific prompts $P$ with the frozen pre-trained parameters: 
\begin{equation}
    \min_{P,W}\, \, \mathbb{E}_{(\boldsymbol{x},\boldsymbol{y})\in\mathcal{D}^k}[\ell(W^\intercal\phi_{\theta_{pre}}(\boldsymbol{x};P),\boldsymbol{y})],
\end{equation}
where $\phi(\boldsymbol{x};P)$ denotes the prompted features by pre-pending the prompts. As a result, with a proper prompt selection mechanism, it can adapt to different downstream tasks. On the other hand, \textit{representation-based} methods only employ the strong representation ability of PTMs, like the nearest class mean (NCM) mechanism \cite{janson2022simple,pelosin2022simpler}. It also freezes the pre-trained weights and extracts the prototype $\boldsymbol{c}$ of each class \cite{janson2022simple}: 
\begin{equation}
    \label{eq:3}
    \boldsymbol{c}_j=\frac{1}{M}\sum_{(\boldsymbol{x},\boldsymbol{y})\in\mathcal{D}^k}\phi_{\theta_{pre}}(\boldsymbol{x}|\boldsymbol{y}=j),
\end{equation}
where $M$ denotes the total number of examples in class $j$. During the inference stage, we first obtain the embedding for each test example and calculate the cosine similarity with each prototype for classification. 
Existing \textit{representation-based} SOTA methods (\textit{i.e.}, ADAM \cite{zhou2023revisiting} and RanPAC \cite{ranpac}) propose variants in this form to achieve better performance. 

\begin{figure*}[!t]
	\begin{center}
 \hspace{-35pt}
            \begin{subfigure}{0.26\linewidth}
    		\centering
    		\includegraphics[width=1\textwidth]{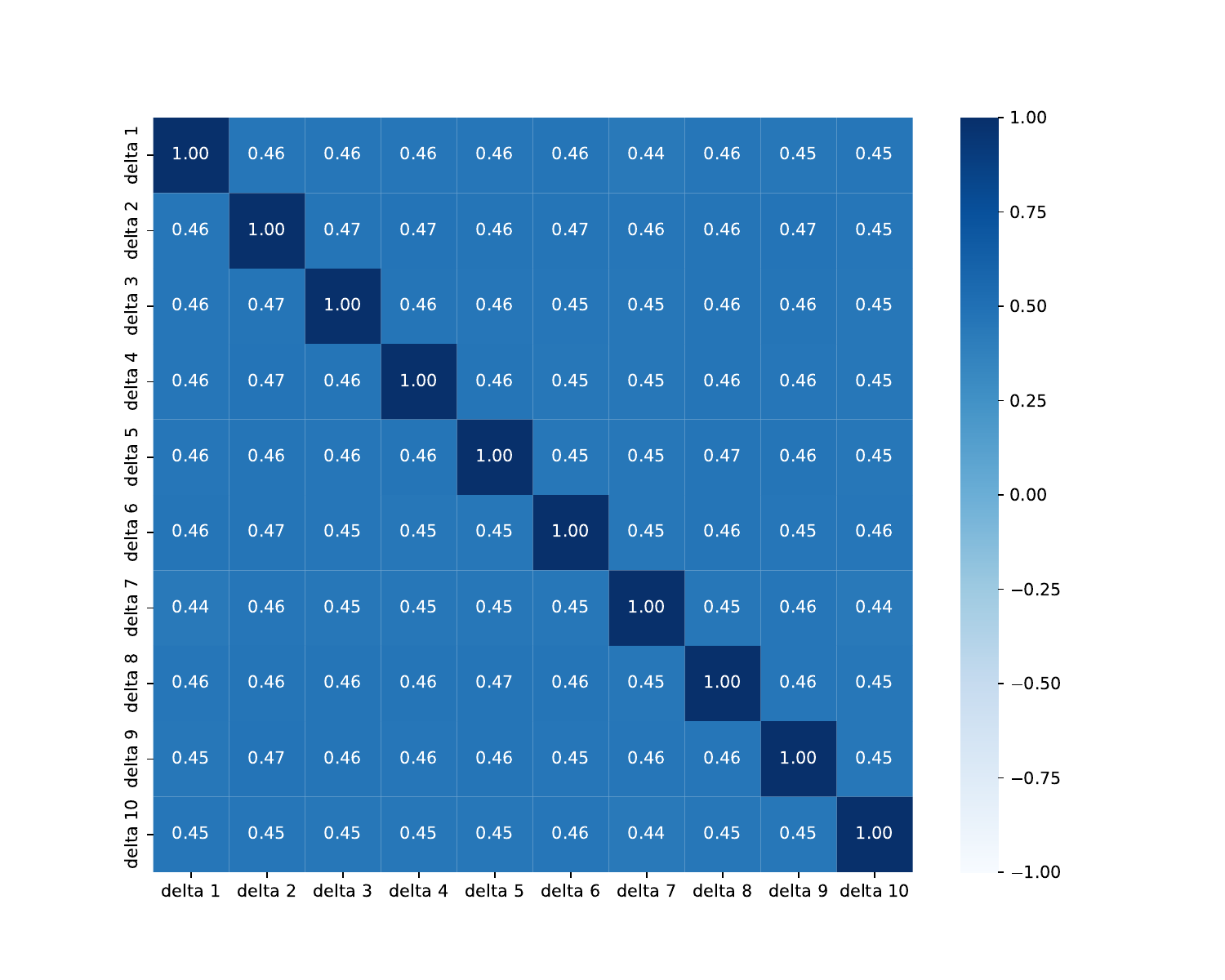}\\
                \vspace{-0.3cm}
    		\caption{Masking 10\%}
    		\label{fig.a}
    	\end{subfigure}
     \hspace{-35pt}
    	\begin{subfigure}{0.26\linewidth}
    		\centering
    		\includegraphics[width=1\textwidth]{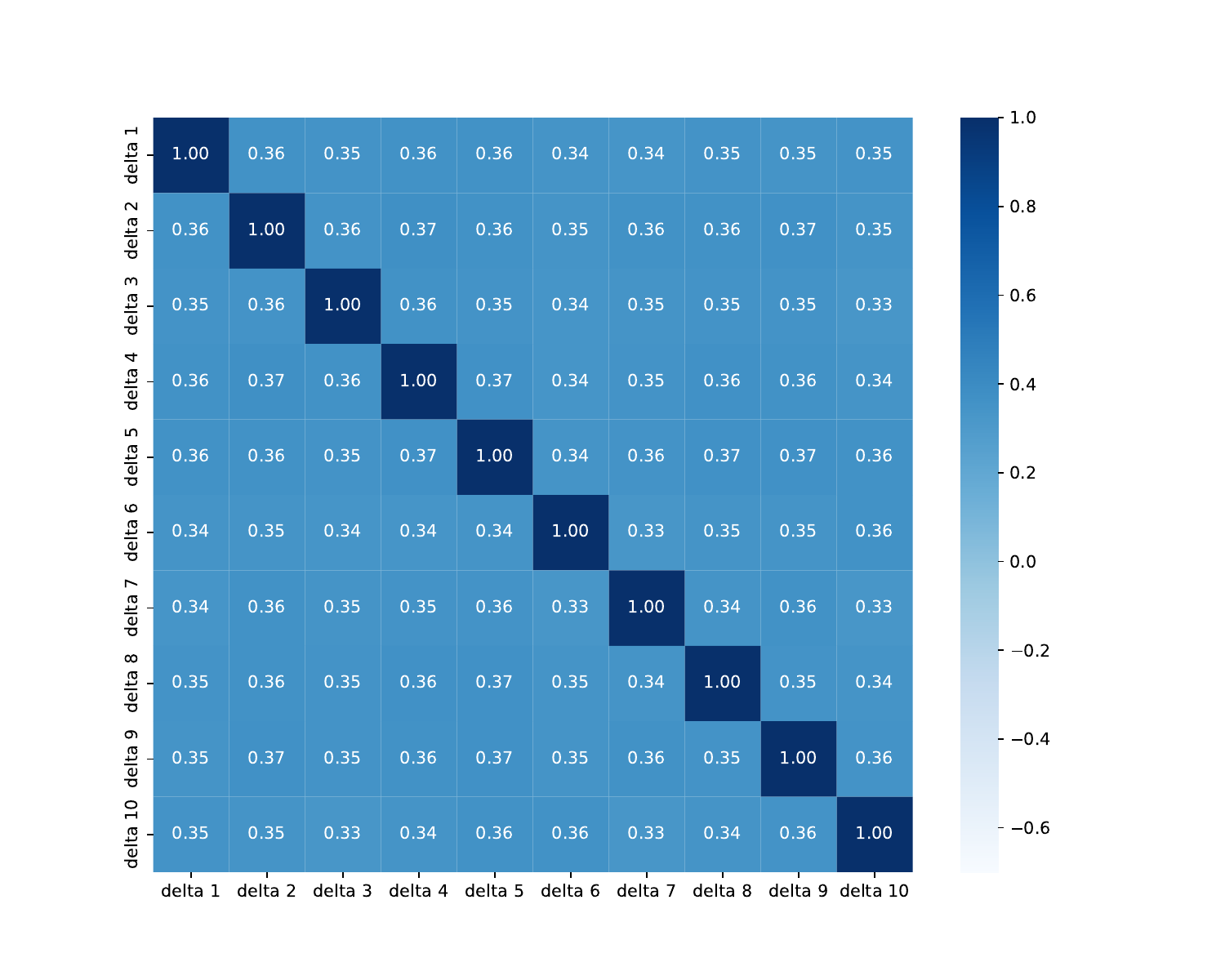}
                \vspace{-0.7cm}
                \caption{Masking 30\%}
    		\label{fig.c}
    	\end{subfigure}
     \hspace{-35pt}
    	\begin{subfigure}{0.26\linewidth}
    		\centering
    		\includegraphics[width=1\textwidth]{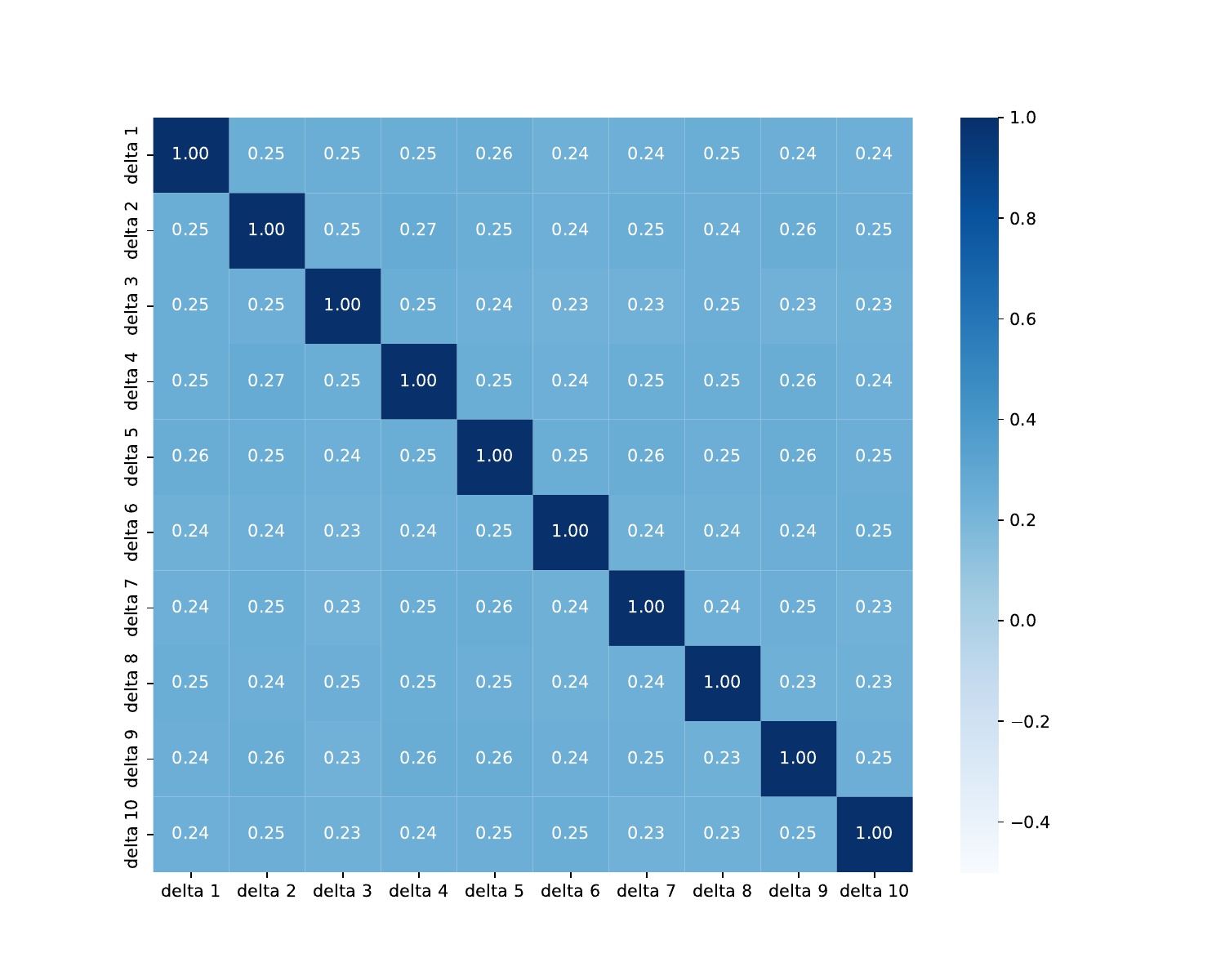}
                \vspace{-0.7cm}
          	\caption{Masking 50\%}
    		\label{fig.e}
    	\end{subfigure}
     \hspace{-35pt}
            \begin{subfigure}{0.26\linewidth}
    		\centering
    		\includegraphics[width=1\textwidth]{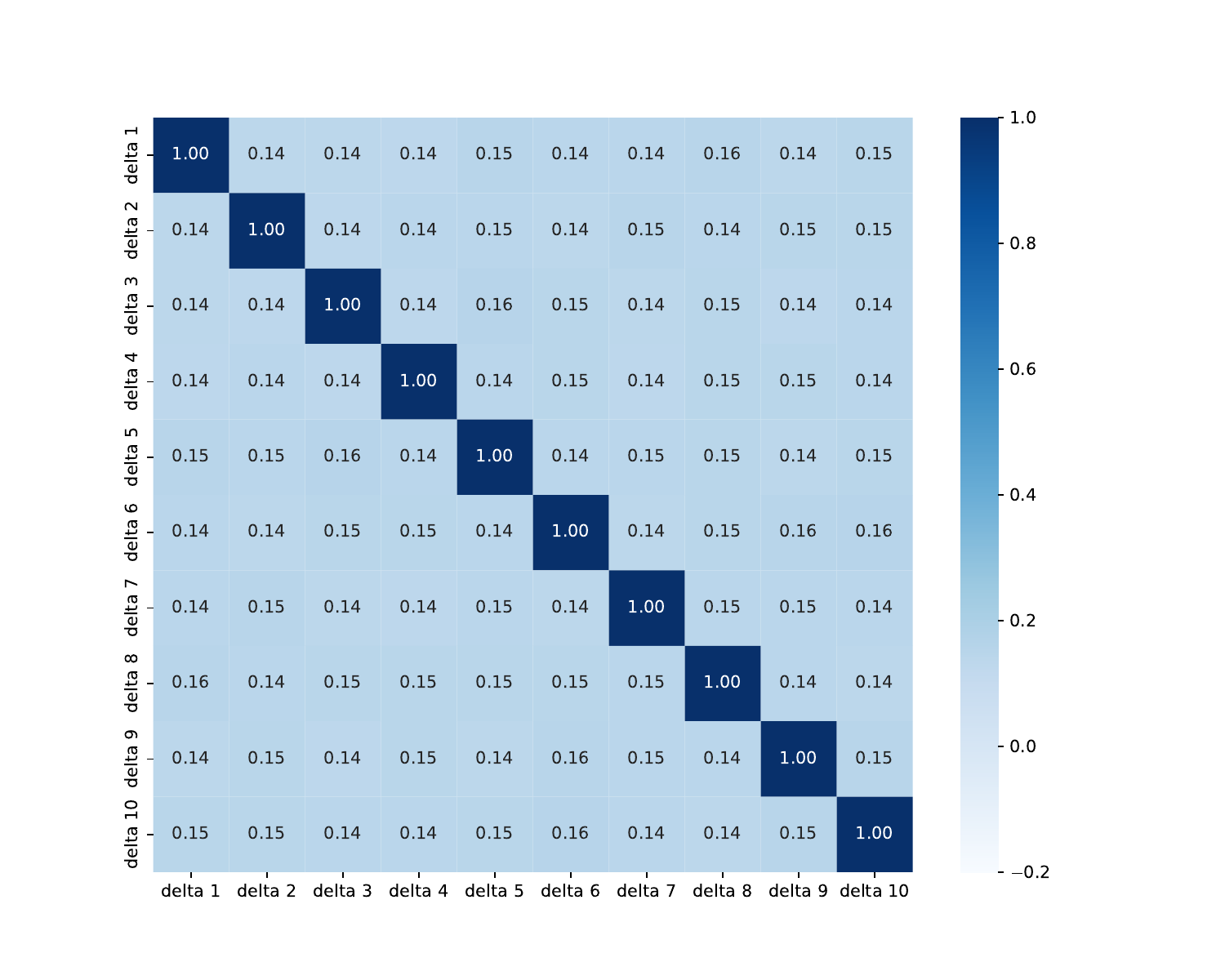}\\
                \vspace{-0.3cm}
    		\caption{Masking 70\%}
    		\label{fig.g}
    	\end{subfigure}
     \hspace{-35pt}
    	\begin{subfigure}{0.26\linewidth}
    		\centering
    		\includegraphics[width=1\textwidth]{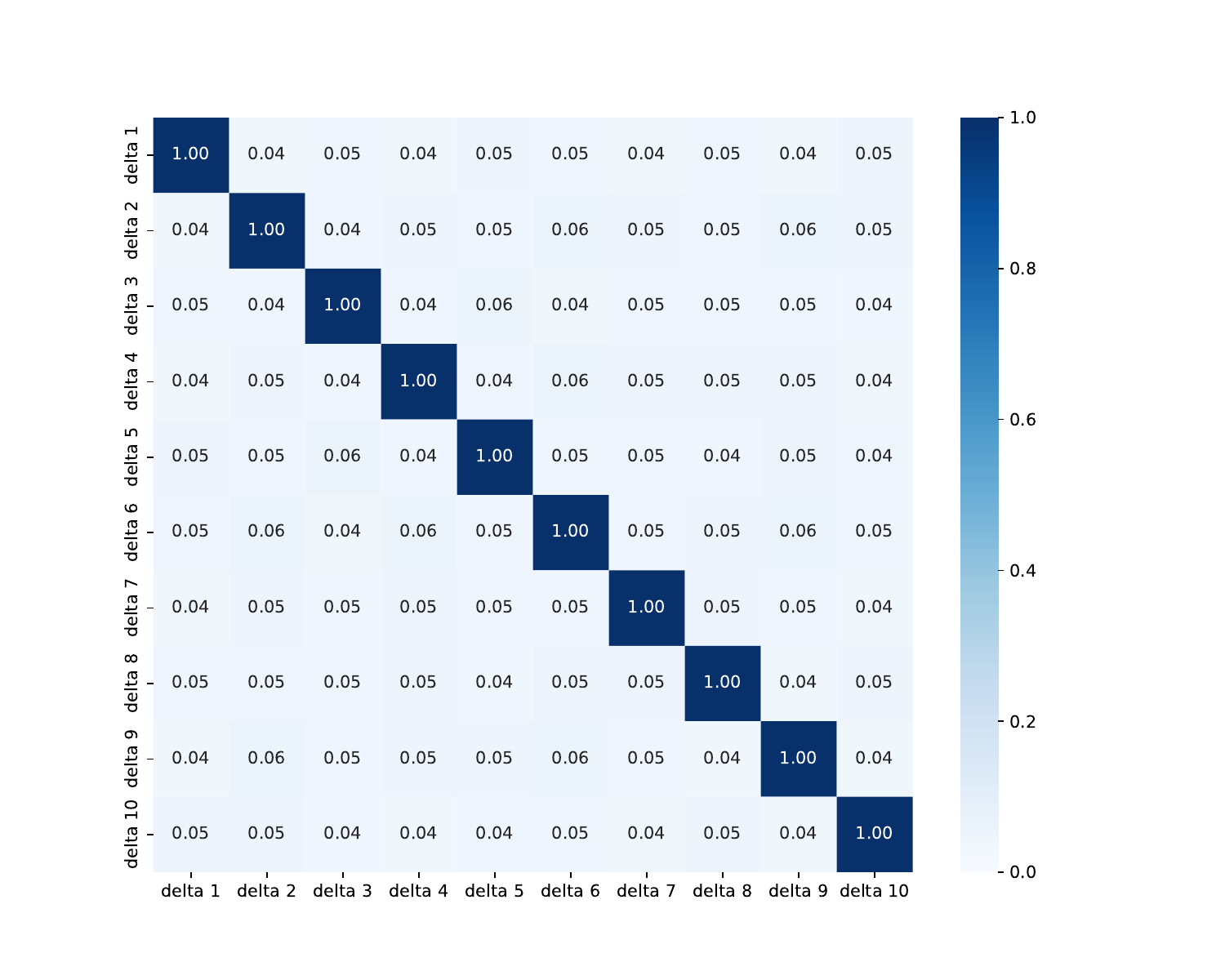}
                \vspace{-0.7cm}
    		\caption{Masking 90\%}
    		\label{fig.i}
    	\end{subfigure}
     \hspace{-35pt}
     
            \caption{The cosine similarity from $\Delta\hat{\theta}^1$ to $\Delta\hat{\theta}^{10}$ with 10\% to 90\% masking rates in CIFAR100 for 10 tasks, where the values in the $i$-th row and $j$-th column represent the similarity between $\Delta\hat{\theta}^i$ and $\Delta\hat{\theta}^{j}$. We demonstrate that high-sparsity deltas (randomly masking 90\% delta parameters in Figure \ref{fig.i}) are more orthogonal to each other. }
            \vspace{-0.3cm}
		\label{fig:o}
	\end{center}
\end{figure*}

\subsection{The Proposed Approach}


As mentioned above, merging sparse orthogonal delta parameters holds enormous promise in mitigating catastrophic forgetting problems in continual learning. Inspired by this, we proposed SoTU (\textbf{S}parse \textbf{O}rthogonal Parameters \textbf{TU}ning), a simple yet effective approach that encourages sparsity in updating parameters to decouple continual deltas and merge each other, while preventing the model from forgetting previously learned knowledge when adapting to new tasks. As shown in Figure \ref{fig:2}, the proposed SoTU approach includes the following three stages. 

\textit{i) Fine-Tuning Stage.} Following the previous works \cite{l2p,dualprompt,codaprompt,ranpac,zhou2023revisiting}, the proposed SoTU approach also utilizes the Vision Transformer (ViT) to initialize the feature extractor ($\theta=\theta_{pre}$). Facing the continual task $\mathcal{D}^k$, we fine-tune the task-specific model as follows, 
\begin{equation}
\theta_{ft}^k, W^k=\mathop{\arg\min}_{\theta,W}\, \, \mathbb{E}_{(\boldsymbol{x},\boldsymbol{y})\sim\mathcal{D}^k}[\ell(W^{\intercal}\phi_{\theta}(\boldsymbol{x}),\boldsymbol{y})].
\end{equation}
where $\theta_{ft}^k$ is the parameter of the learned feature extractor $\phi_{\theta_{ft}^k}(\cdot)$, and $W^k$ denotes the classification layer of task $k$. 
Note that $W^k$ will not participate in the following processes, and its role is only to assist in learning the feature representation $\theta_{ft}^k$ of the current task $k$.

\textit{ii) Delta Masking Stage.} In this stage, we calculate the delta parameters ($\Delta\theta$) for each task and randomly mask the delta based on the Bernoulli distribution. 
\begin{equation}
    \Delta\theta^k=\theta_{ft}^k-\theta_{pre},
\end{equation}
\begin{equation}
    \Delta\hat{\theta}^k=\Delta\theta^k\odot\boldsymbol{m}. 
\end{equation}
where $\boldsymbol{m}\sim Bernoulli(p)$, and $p$ is the masking probability. To maintain the high sparsity (orthogonality) of delta $\Delta\hat{\theta}$, we set $p=0.9$ in our experiments. 
It is worth noting that we only need to store the delta parameter $\Delta\hat{\theta}^k$ for each task $k$. This storage cost is low, as there are only 10\% delta parameters.

\textit{iii) Delta Merging Stage.} After that, we could obtain the final model $\theta^k$ by merging those \textit{sparse orthogonal} deltas for all seen tasks without forgetting, 
\begin{equation}
   \label{eq:7} \theta^k=\theta_{pre}+\Delta\hat{\theta}^1+\Delta\hat{\theta}^2+...+\Delta\hat{\theta}^k. 
\end{equation}
As mentioned above, these high-sparsity deltas $\Delta\hat{\theta}$ can store task-specific knowledge from multiple domains, while merging them into one feature extractor ($\theta^k$) can perform well in tasks $1$ to $k$. The detailed discussion is as follows.

\textbf{Sparse Orthogonal Makes Better Performance.} We demonstrate the cosine similarity from $\Delta\hat{\theta}^1$ to $\Delta\hat{\theta}^{10}$ in Figure \ref{fig:o}. The values in the $i$-th row and $j$-th column represent the similarity between $\Delta\hat{\theta}^i$ and $\Delta\hat{\theta}^{j}$, with a diagonal value of 1. The results show that high-sparsity deltas (Figure \ref{fig.i}) are more orthogonal to each other. In other words, when we merge these high-sparsity deltas as shown in Eq.\ref{eq:7}, there will be fewer parameter conflicts due to the large number of $0$ elements, while leading to better performance in all seen tasks (as mentioned above in Figure \ref{fig:1}). 


\textbf{Discussion of Continual Learning in Feature Space.} It is worth highlighting that our SoTU method is capable of achieving optimal feature representation for streaming data without the need for any elaborate classifier designs. We focus more on the effectiveness of the SoTU approach in combating catastrophic forgetting in the feature space. We believe this is a more flexible approach that can be applied to more real-world tasks. Therefore, it can also serve as a Plug-and-Play method. In this paper, we mainly validate the effectiveness of SoTU in the most popular Class-Incremental Learning (CIL) problems, where the task ID is unknown and the class number is changing. We employ the Nonlinear Random Projection mechanism \cite{ranpac,schmidt1992feed,mcdonnell2016importance,chen1996rapid} to transform the learned mixture feature $\phi_{\theta^k}$ into the linear separability space for better class-prototype-based classification (Eq.\ref{eq:3}).

\begin{table*}[t]
    \centering
    \resizebox{\textwidth}{!}{\begin{tabular}{c|c c c c c c c c c c c c}
        \toprule
        \multirow{2}{*}{Method}& \multicolumn{2}{c}{CIFAR100} & \multicolumn{2}{c}{ImageNet-R} & \multicolumn{2}{c}{ImageNet-A} & \multicolumn{2}{c}{CUB200} & \multicolumn{2}{c}{VTAB} & \multicolumn{2}{c}{Cars196}  \\
        & $\Bar{\mathcal{A}}$ & $\mathcal{A}_T$ & $\Bar{\mathcal{A}}$ & $\mathcal{A}_T$ & $\Bar{\mathcal{A}}$ & $\mathcal{A}_T$ & $\Bar{\mathcal{A}}$ & $\mathcal{A}_T$ & $\Bar{\mathcal{A}}$ & $\mathcal{A}_T$ & $\Bar{\mathcal{A}}$ & $\mathcal{A}_T$ \\
        \hline
        Joint full tuning & - & 93.8\% & - & 86.6\% & - & 70.8\% & - & 90.5\% & - & 92.0\% & - & 86.9\%\\
        \hline
        \rowcolor{gray!20}
        iCaRL \cite{rebuffi2017icarl} & 89.7\% & 79.3\% & 72.4\% & 58.8\% & 39.7\% & 29.5\% & 87.6\% & 80.5\% & 74.0\% & 85.9\% & 74.7\% & 62.4\%\\
        \rowcolor{gray!20}
        DER \cite{yan2021dynamically} & 88.8\% & 79.8\% & 82.1\% & 76.3\% & 39.7\% & 29.1\% & 88.9\% & 84.7\% & 93.1\% & 89.8\% & 75.2\% & 69.9\%\\
        \rowcolor{gray!20}
        FOSTER \cite{wang2022foster} & 91.8\% & 87.8\% & 82.6\% & 76.4\% & 46.5\% & 37.7\% & 81.8\% & 76.7\% & 87.9\% & 80.2\% & 44.3\% & 42.9\%\\
        \rowcolor{gray!20}
        MEMO \cite{zhou2022model} & 87.1\% & 77.2\% & 74.4\% & 65.7\% & 39.2\% & 24.9\% & 88.9\% & 82.7\% & 81.5\% & 86.9\% & 70.6\% & 60.5\% \\
        \hline
        \rowcolor{green!20}
        L2P \cite{l2p} & 89.3\% & 84.4\% & 78.1\% & 72.7\% & 53.9\% & 44.9\% & 79.6\% & 68.1\% & 83.4\% & 61.7\% & 58.2\% & 45.3\% \\
        \rowcolor{green!20}
        DualPrompt \cite{dualprompt} & 87.4\% & 82.4\% & 75.1\% & 69.1\% & 57.2\% & 47.3\% & 79.8\% & 68.7\% & 86.2\% & 73.6\% & 53.4\% & 37.7\% \\
        \rowcolor{green!20}
        CODA-Prompt \cite{codaprompt} & 91.3\% & 86.9\% & 78.5\% & 73.4\% & 63.9\% & 52.7\% & 84.1\% & 79.3\% & 88.6\% & 89.3\% & 52.1\% & 45.4\% \\
        \hline
        \rowcolor{yellow}
        HiDe-Prompt \cite{wang2024hierarchical} & 92.8\% & 90.6\% & 80.5\% & 76.7\% & 57.5\% & 43.2\% & 87.8\% & 87.2\% & 84.8\% & 84.6\% & 53.6\% & 52.0\% \\
        \rowcolor{yellow}
        ESN \cite{wang2023isolation} & 90.1\% & 86.1\% & 79.6\% & 74.4\% & 56.7\% & 44.9\% & 82.3\% & 78.4\% & 85.9\% & 85.2\% & 72.8\% & 56.9\% \\
        \hline
        \rowcolor{red!20}
        SimpleCIL \cite{janson2022simple} & 82.2\% & 76.2\% & 67.1\% & 61.3\% & 59.4\% & 49.3\% & 90.6\% & 85.2\% & 90.9\% & 83.6\% & 50.4\% & 37.8\% \\ 
        \rowcolor{red!20}
        ADAM (SSF) \cite{zhou2023revisiting} & 83.9\% & 89.5\% & 79.1\% & 72.2\% & 62.4\% & 52.1\% & 90.5\% & 85.6\% & 91.6\% & 84.3\% & 52.8\% & 40.5\% \\
        \rowcolor{red!20}
        ADAM (Ada) \cite{zhou2023revisiting} & 90.0\% & 85.8\% & 79.1\% & 72.7\% & 66.2\% & 56.4\% & 91.1\% & 86.2\% & 90.9\% & 83.6\% & 56.0\% & 43.2\% \\
        \rowcolor{red!20}
        RanPAC \cite{ranpac} & 94.0\% & 90.8\% & 83.2\% & 77.9\% & 72.7\% & 62.4\% & 92.6\% & 88.9\% & 95.3\% & 92.2\% & 82.8\% & 75.1\% \\
        \hline
        \textbf{SoTU (Ours)} & \textbf{94.5\%} & \textbf{91.2\%} & \textbf{84.4\%} & \textbf{79.5\%} & \textbf{75.1\%} & \textbf{64.7\%} & \textbf{92.7\%} & \textbf{89.1\%} & \textbf{96.7\%} & \textbf{93.4\%} & \textbf{85.3\%} & \textbf{78.8\%}\\
        \bottomrule
    \end{tabular}}
    \vspace{-0.3cm}
    \caption{Average and final accuracy on six datasets, each dataset is divided into 10 tasks. Different colors (\colorbox{gray!20}{\rule{0pt}{0.1cm}\rule{0.1cm}{0pt}}, \colorbox{green!20}{\rule{0pt}{0.1cm}\rule{0.1cm}{0pt}}, \colorbox{red!20}{\rule{0pt}{0.1cm}\rule{0.1cm}{0pt}}, \colorbox{yellow}{\rule{0pt}{0.1cm}\rule{0.1cm}{0pt}}) indicate rehearsal, prompt-based, and representation-based, model merge methods, respectively.}
    \label{table:1}
\end{table*}

\begin{figure*}[!t]
	\begin{center}
            \begin{subfigure}{0.32\linewidth}
    		\centering
    		\includegraphics[width=1\textwidth]{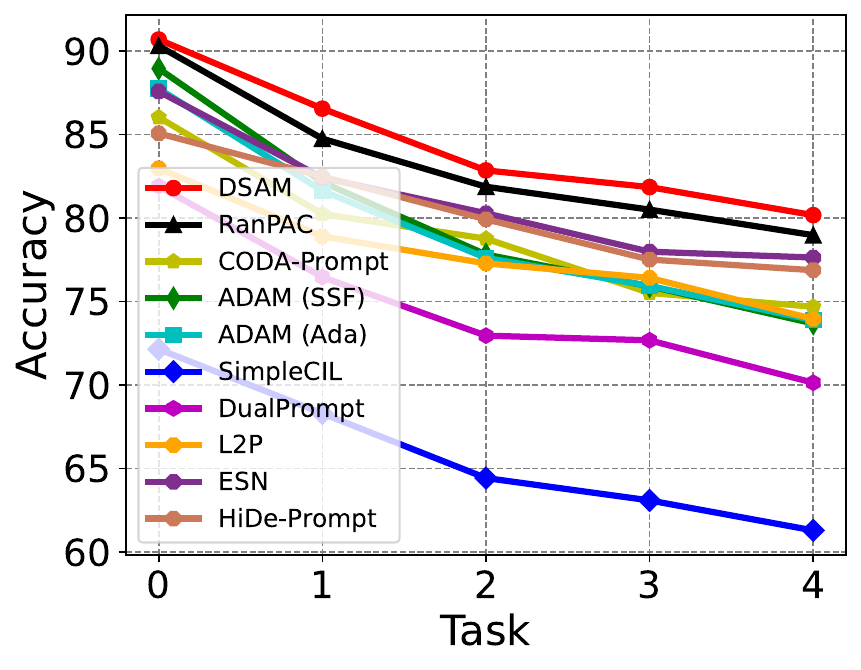}
                \vspace{-0.3cm}
    		\caption{ImageNet-R}
    		\label{fig.add_examples_a}
    	\end{subfigure}
    	\begin{subfigure}{0.32\linewidth}
    		\centering
    		\includegraphics[width=1\textwidth]{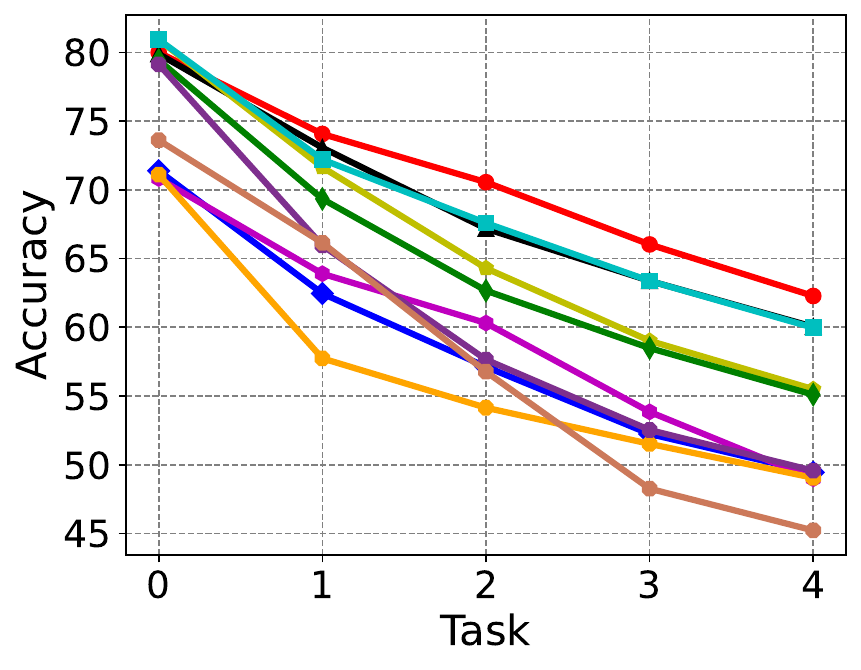}
                \vspace{-0.3cm}
          	\caption{ImageNet-A}
    		\label{fig.remove_examples_noisy}
    	\end{subfigure}
    	\begin{subfigure}{0.32\linewidth}
    		\centering
    		\includegraphics[width=1\textwidth]{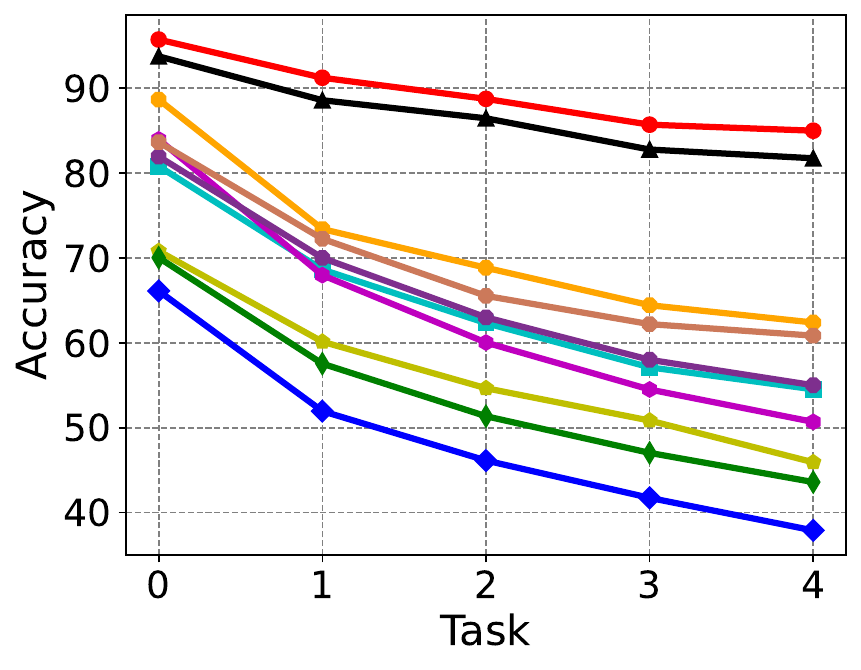}
                \vspace{-0.3cm}
    		\caption{Cars196}
    		\label{fig.remove_examples_clean}
    	\end{subfigure}
            \caption{Continual learning performance comparison with PTM-based methods on three difficult datasets, \textit{i.e.}, ImageNet-R, ImageNet-A, and Cars196. Each dataset is divided into 5 tasks.}
            \vspace{-0.3cm}
		\label{fig:3}
	\end{center}
\end{figure*}

\subsection{Theoretical Analysis}
\label{sec:3.3}

In this subsection, we provide theoretical explanations for why high-sparsity delta can store task-specific knowledge and preserve performance similar to that of pre-training.


\begin{theorem}
\label{theorem:attention}
For a specific fine-tuned attention block parameterized by $\boldsymbol{W} = [\boldsymbol{W}_K, \boldsymbol{W}_Q, \boldsymbol{W}_V]$.  The perturbed attention $\hat{A}_{ij}$ are bounded within $\left[\frac{1+\delta^{min}}{1+\delta^{max}}A_{ij}, \frac{1+\delta^{max}}{1+\delta^{min}}A_{ij}\right]$, when $\boldsymbol{W}$ is perturbed with small $\Delta\boldsymbol{W} = [\Delta\boldsymbol{W}_K, \Delta\boldsymbol{W}_Q, \Delta\boldsymbol{W}_V]$.
\end{theorem}

The proof can be found in the supplementary material.
Theorem.\ref{theorem:attention} is based on the small $\delta$ hypothesis, and its supporting results can also be found in the supplementary material. Theorem.\ref{theorem:attention} implies that randomly masking the delta parameter $\Delta\theta$ can retain its attention map and achieve competitive performance on corresponding tasks. The visualization results of the attention map also validate this.

\section{Experiments}
\label{sec:4}

\textbf{Datasets.} Following \cite{ranpac,zhou2024continual}, we evaluate the performance on CIFAR100 \cite{krizhevsky2009learning}, ImageNet-R \cite{hendrycks2021many}, ImageNet-A \cite{hendrycks2021natural}, CUB200 \cite{wah2011caltech}, VTAB \cite{zhai2019large} and Cars196 \cite{zhang2023slca}. Apart from CIFAR100 and CUB200, the other four datasets are acknowledged to have a large domain gap with ImageNet, making the PTM less generalizable and increasing the difficulty of CL. Following \cite{ranpac}, we split each dataset into 5 ($T=5$) and 10 ($T=10$) tasks on average. Before splitting, we randomly shuffle all classes with the same random seed for a fair comparison.

\begin{table*}[t]
    \centering
        \resizebox{\textwidth}{!}{\begin{tabular}{c| c | c c c c c c }
            \toprule
            \multirow{2}{*}{Method}& \multirow{2}{*}{Buffer Size}& \multicolumn{2}{c}{CIFAR100} & \multicolumn{2}{c}{ImageNet-R} & \multicolumn{2}{c}{CUB200}\\
            & & $\Bar{\mathcal{A}}$ & $\mathcal{A}_T$ & $\Bar{\mathcal{A}}$ & $\mathcal{A}_T$ & $\Bar{\mathcal{A}}$ & $\mathcal{A}_T$\\
            \hline
            RanPAC w/o Nonlinear Projection & 10 / cls & 81.9\% & 75.4\% & 58.7\% & 52.2\% & 81.7\% & 74.6\%\\
            & 500 / cls & 82.2\% & 75.9\% & 60.2\% & 53.9\% & 84.3\% & 77.3\%\\
            SoTU w/o Nonlinear Projection & 10 / cls & 89.2\% (\textbf{+7.3\%}) & 83.8\% (\textbf{+8.4\%}) & 72.7\% (\textbf{+14.0\%}) & 67.9\% (\textbf{+12.7\%}) & 84.7\% (\textbf{+3.0\%}) & 78.8\% (\textbf{+4.2\%})\\
            & 500 / cls & 89.5\% (\textbf{+7.3\%}) & 84.8\% (\textbf{+8.9\%}) & 73.9\% (\textbf{+13.7\%}) & 69.9\% (\textbf{+16.0\%}) & 85.7\% (\textbf{+1.4\%}) & 79.8\% (\textbf{+2.6\%})\\
            \hline

        \end{tabular}}
        \vspace{-0.3cm}
        \caption{The average and final performance comparison in feature space. The two methods only employ the learned feature extractor and nearest class mean (NCM) mechanism \cite{janson2022simple,pelosin2022simpler} for prediction. 
        }
        \label{table:2}
\end{table*}

\begin{figure*}[t]
    \centering
    \vspace{-0.3cm}
    \includegraphics[width=0.95\textwidth]{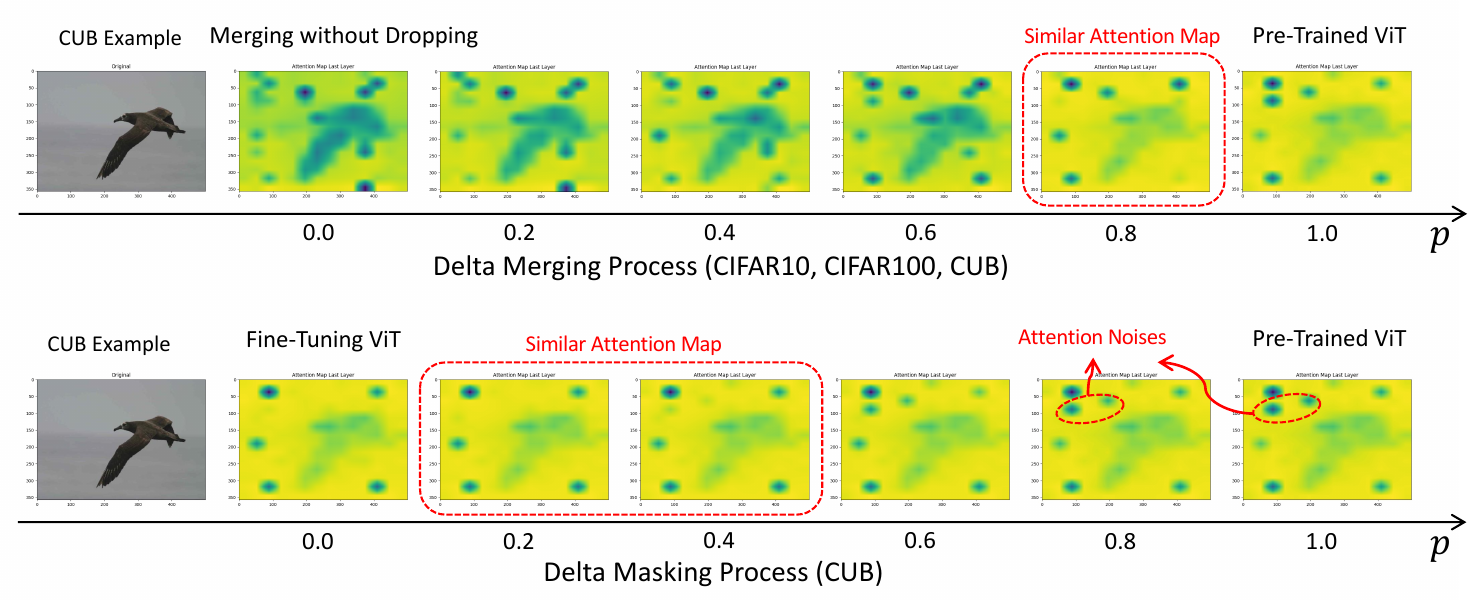}
    \caption{The visualization of attention map in delta masking and merging processes. Randomly masking delta parameters can retain its attention map. Directly merging deltas without masking will cause serious parameter collision, damaging the attention map. Merging high-sparsity deltas can achieve similar attention maps, as they are orthogonal to each other. }
    \label{fig:4}
\end{figure*}

\textbf{Baselines.} To validate the effectiveness of the proposed SoTU approach, we compare the following twelve methods from four categories. 
\begin{itemize}
    \item 
    \textcolor{gray}{Rehearsal methods:} \textit{i)} \textbf{iCaRL} \cite{rebuffi2017icarl}: it stored a subset of exemplars per class and selected to best approximate class means in the learned feature space. \textit{ii)} \textbf{DER} \cite{yan2021dynamically}: it utilizes a dynamically expandable representation for more effective incremental concept modeling. \textit{iii)} \textbf{FOSTER} \cite{wang2022foster}: it dynamically expands new modules and removes redundant parameters and feature dimensions to maintain the single backbone model. \textit{iv)} \textbf{MEMO} \cite{zhou2022model}: it extends specialized layers based on the shared generalized representations, efficiently extracting diverse representations with modest cost and maintaining representative exemplars. Note that all rehearsal methods store 20 examples per class in our experiments.
    \item 
    \textcolor{green!70}{Prompt-based methods:} \textit{v)} \textbf{L2P} \cite{l2p}: it incorporates visual prompt tuning into CIL using a pre-trained ViT and establishes a prompt pool for selecting the instance-specific prompts. \textit{vi)} \textbf{DualPrompt} \cite{dualprompt}: it proposes general and expert prompts based on L2P. \textit{vii)} \textbf{CODA-Prompt} \cite{codaprompt}: it improves the prompt selection process with an attention mechanism.
    \item 
    \textcolor{red!70}{Representation-based methods:} \textit{viii)} \textbf{SimpleCIL} \cite{janson2022simple}: it sets the classifiers of the pre-trained model to prototype features without any training. \textit{ix)} \textbf{ADAM} \cite{zhou2023revisiting}: it employs parameter-efficient tuning to acquire an adapted model, and we report its two versions in this paper. \textit{x)} \textbf{RanPAC} \cite{ranpac}: it uses an online LDA classifier to remove classwise correlations for better separability while constructing a random projection layer for prediction.
    \item 
    \textcolor{orange}{Model Mixture-based methods:} \textit{xi)} \textbf{ESN} \cite{wang2023isolation}: it initializes and trains a new classifier head when facing a new task, while designing a voting strategy for these classifier heads during the inference stage. \textit{xii)} \textbf{HiDe-Prompt} \cite{wang2024hierarchical}: it applies a prompt merge after each continual learning stage.
\end{itemize}
Moreover, we also report the results of ``\textbf{Joint full tuning}'', which denotes the continual learning upper bound by full fine-tuning model on the entirety of the dataset.
It is worth noting that all baselines (including rehearsal methods) employ the ViT pre-trained model on ImageNet21K, \textit{i.e.}, ViT-B/16-IN21K. We implemented all compared methods by PILOT \cite{sun2023pilot} open-source code.

\textbf{Evaluation Metrics.} We follow \cite{lopez2017gradient} and use two standard CL metrics, Average Accuracy $\Bar{\mathcal{A}}$ and Final Accuracy $\mathcal{A}_T$, which are defined as $\Bar{\mathcal{A}}=\frac{1}{T}\sum_{k=1}^T R_k$ and $\mathcal{A}_T=R_T$, where $R_k$ is the classification accuracy from task $1$ to task $k$. 


\subsection{Main Results}

In this subsection, we compared the proposed SoTU approach with twelve CL baselines on six commonly used benchmarks. 

\textbf{Performance Comparison with SOTA Methods.} We extensively evaluate the SoTU approach by comparing it with various baselines across multiple datasets and metrics. The average results of 10 tasks are demonstrated in Table \ref{table:1}. It can be observed that no matter which dataset or metric is used, the proposed SoTU approach always outperforms other methods in all cases. 
These rehearsal methods, storing 20 examples per class, can achieve competitive results on many tasks, except ImageNet-A. Preserving some exemplars can effectively revisit the learned knowledge. Compared with it, other baselines do not store any examples during the CL process. The prompt-based approaches, which attempt to learn task-specific prompts, perform well on typical CL benchmarks (\textit{e.g.}, CIFAR100, VTAB), which however fail on some difficult tasks (\textit{e.g.}, Imagenet-A and Cars196). The larger domain gaps between these datasets and the original ImageNet data further examine the effectiveness of existing PTM-based methods in combating catastrophic forgetting. Compared with the current SOTA method (RanPAC), the proposed SoTU approach also shows greater superiority in these more challenging datasets, \textit{i.e.}, achieving final accuracy improvements of $\bm{+3.7\%}$ in Cars196, $\bm{+2.3\%}$ in ImageNet-A, and $\bm{+1.6\%}$ in ImageNet-R. 

\textbf{Comparison of Continual Learning Process.} To further validate the effectiveness of the proposed approach, we compared the PTM-based methods in three more difficult datasets, \textit{i.e.}, ImageNet-R, ImageNet-A, and Cars196. Each dataset is divided into 5 tasks. The results in Figure \ref{fig:3} consistently validate that the proposed SoTU can better tackle catastrophic forgetting problems during the CL process. SoTU can achieve better accuracy performance in almost all cases. 


\begin{figure*}[!t]
	\begin{center}
    	\begin{subfigure}{0.32\linewidth}
    		\centering
    		\includegraphics[width=1\textwidth]{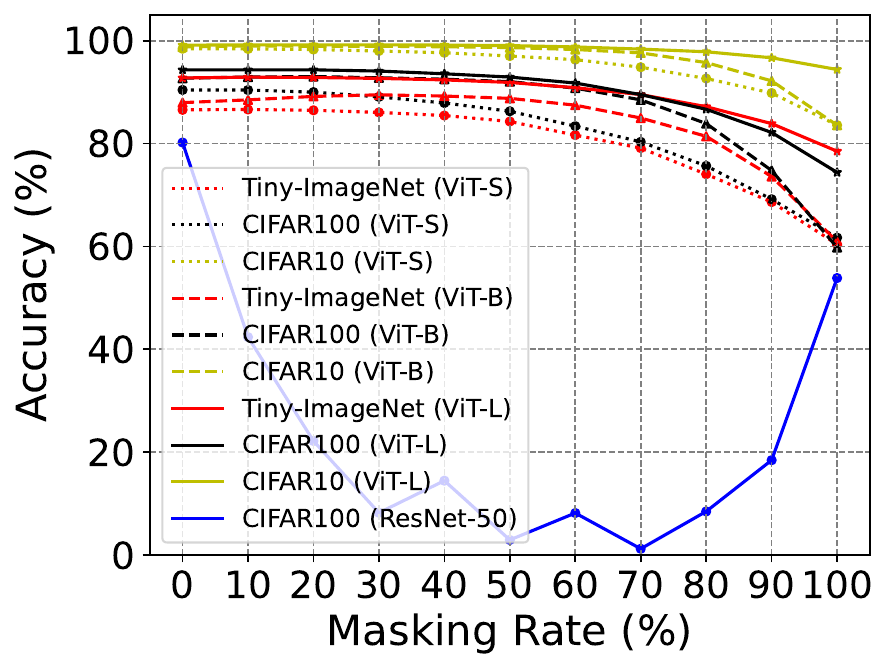}
          	\caption{Delta Masking}
    		\label{fig.delta_sampling}
    	\end{subfigure}
            \begin{subfigure}{0.32\linewidth}
    		\centering
    		\includegraphics[width=1\textwidth]{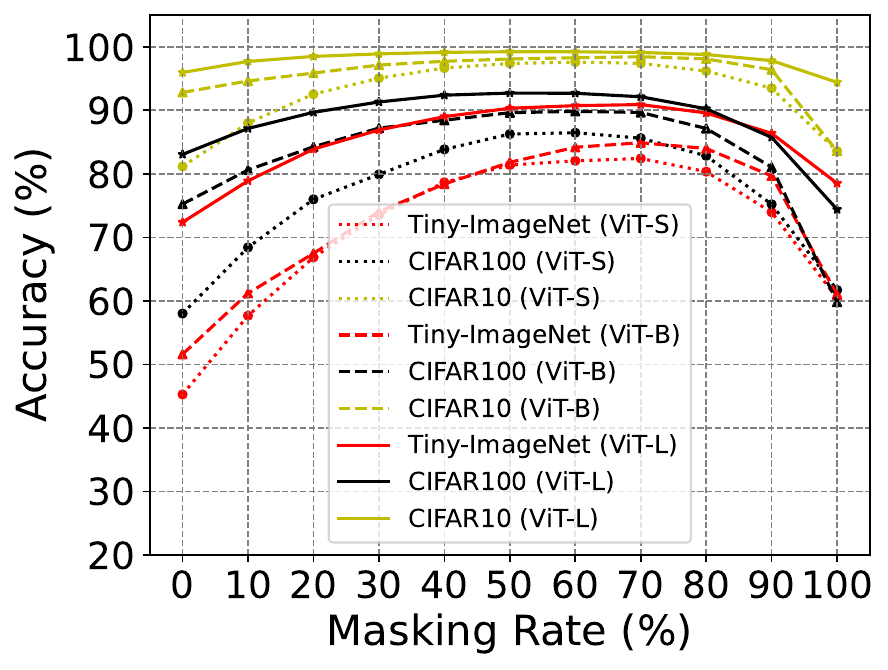}\\
    		\caption{Delta Merging}
    		\label{fig.delta_merging}
    	\end{subfigure}
            \begin{subfigure}{0.32\linewidth}
    		\centering
    		\includegraphics[width=1\textwidth]{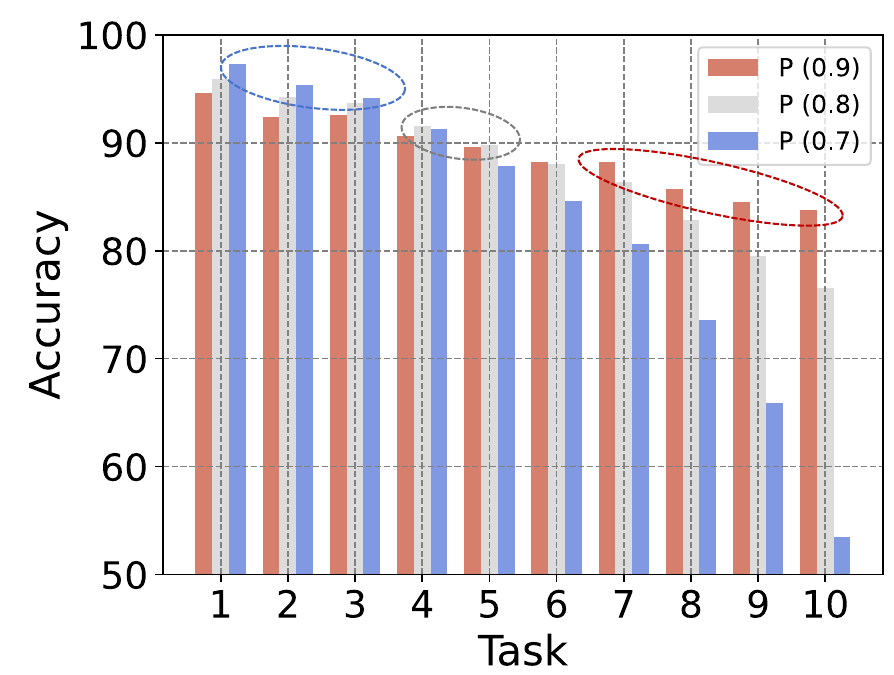}\\
    		\caption{Analyze Masking Rate}
    		\label{fig.drop}
    	\end{subfigure}
            \vspace{-0.1cm}
            \caption{The characteristics of delta masking and merging.}
            \vspace{-0.4cm}
		\label{fig:5}
	\end{center}
\end{figure*}

\subsection{Study Catastrophic Forgetting in Feature Space}
In this subsection, we aim to study the effectiveness of the proposed SoTU approach in combating catastrophic forgetting in feature space. Specifically, we follow the nearest class mean (NCM) mechanism \cite{janson2022simple,pelosin2022simpler} by removing the classification head $W$ and only use the learned feature $\phi_{\theta}(\cdot)$ for prediction (as shown in Eq \ref{eq:3}). 

\textbf{The Effectiveness of SoTU in Feature Space.} We perform the experiments on three datasets with different buffer sizes for 10 tasks, where the buffer size denotes the number of examples to calculate the prototypes for each class. As shown in Table \ref{table:2}, without nonlinear feature projection, the proposed SoTU approach can significantly improve classification accuracy than the RanPAC method in all cases, especially on the ImageNet-R dataset. These results demonstrate that the learned mixture feature by our SoTU approach can better combat catastrophic forgetting problems in feature space. 

\textbf{Visualize Attention Map of SoTU.}
To further study the effectiveness of SoTU, we visualize the attention map of the UCB example during the delta masking and merging processes with different masking rates. Note that the delta masking process is conducted on the CUB200 dataset, and the delta merging process is conducted on three datasets, \textit{i.e.}, CIFAR10, CIFAR100, and CUB200. 
As shown in Figure \ref{fig:4}, it can be observed that even if randomly masking $40\%\sim 60\%$ delta parameters, the model can still obtain similar attention maps (red dashed box), which is also validated in our theoretical analysis. However, masking more delta parameters may generate attention noises (red dashed ellipse) while hurting the model's performance. \textbf{\textit{Most importantly, in the delta merging process, directly merging three fine-tuned ViTs without masking deltas will cause serious parameter collision, damaging the attention map.}} With a higher sparsity, merging these orthogonal delta parameters can also achieve similar attention maps (red dashed box), which means that it can perform well for downstream tasks. These results once again validate that the proposed SoTU approach can effectively learn knowledge from multiple domains and obtain better feature representation without forgetting.

\subsection{Study of Delta Masking and Merging}

In this subsection, we aim to study the effect of model size, model architecture, masking probability $p$, and different datasets in delta masking and merging. We thus construct experiments on CIFAR10, CIFAR100, and Tiny-ImageNet with the Vit-S, ViT-B, ViT-L, and ResNet50 models.


\textbf{Study on Delta Masking.} Figure \ref{fig.delta_sampling} demonstrates the performance of delta masking in three tasks with different models and masking rates. Note that the 0 and 100 masking rates denote the fully fine-tuned and pre-trained models, respectively. It can be observed the following three phenomena: 
\textit{i)} Randomly masking $80\%\sim90\%$ delta parameters can still maintain competitive performance. These results reveal a similar \textit{delta sparsity} characteristic of PTM that a few delta parameters are actually enough to store task-specific knowledge, which is also validated in language model \cite{yu2023language} and lottery ticket hypothesis \cite{frankle2018lottery}.
\textit{ii)} The larger parameter quantities can tolerate higher delta sparsity, as masking more deltas has the least impact on ViT-L, followed by ViT-B and finally ViT-S. \textit{iii)} Other model structures, \textit{e.g.}, ResNet50 (blue line), may not have the \textit{delta sparsity} characteristic, as masking its deltas will seriously hurt the model performance. 

\textbf{Study on Delta Merging.} Figure \ref{fig.delta_merging} displays the performance of merging three tasks with different ViT models, and the phenomena can be summarized as follows: \textit{i)} Merging high-sparsity deltas ($p\approx0.7$) can achieve competitive performance with fully fine-tuned models, which occur across different datasets and ViT models. \textit{ii)} Merging low-sparsity deltas will seriously hurt the model performance, possibly because parameter collisions destroy the learned task-specific knowledge, which is also validated in Figure \ref{fig:4}. \textit{iii)} Masking excessive deltas also hurt the model's performance, as task-specific knowledge is missing. 

\textbf{Analysis of the masking rate $p$.} As shown in Figure \ref{fig.drop}, we employed different masking rates ($p=0.9,0.8,0.7$) and performed the SoTU approach in CIFAR100 for 10 ($T=10$) tasks. For convenience, we define the delta-sparsity as $1-p$, which denotes how many delta parameters are retained. We found an interesting phenomenon that the delta-sparsity $1-p$ is approximately $\frac{1}{T}$. As shown in Figure \ref{fig.drop}, when we maintain 30\% delta-sparsity ($p=0.7$), it can achieve better performance in the first 3 tasks (blue dashed ellipse), which however fails in the following 7 tasks. Similar phenomena also occur in the other two cases (grey and red dashed ellipses). These results reveal that when faced with multiple tasks, a better delta masking strategy is to preserve task-specific knowledge as much as possible while reducing parameter collisions between different tasks. 


\section{Conclusion}
In this paper, we reveal the benefit of \textit{sparse orthogonal} parameters for continual learning, \textit{i.e.}, merging high-sparsity delta parameters can effectively learn specific knowledge from multiple downstream tasks without forgetting. We demonstrated that the effectiveness behind this is due to the parameters' orthogonality across multiple tasks. 
Based on this, we propose a novel CL method called SoTU that can achieve optimal feature representation for streaming data without any elaborate classifier designs. Facing continual tasks, it transforms the learned knowledge from multiple domains into the merging of corresponding orthogonal delta parameters. The experimental results on multiple CL benchmarks show the superiority of SoTU. Both theoretical analysis and visualization also validate the effectiveness of SoTU. In the future, we will extend the SoTU approach to other continual learning tasks, \textit{e.g.}, natural language generation.

\bibliography{aaai25}

\end{document}